\def\year{2020}\relax
\documentclass[letterpaper]{article} 
\usepackage{aaai20}  
\usepackage{times}  
\usepackage{helvet} 
\usepackage{courier}  
\usepackage[hyphens]{url}  
\usepackage{graphicx} 
\usepackage{graphicx}  
\usepackage{todonotes}
\usepackage{multirow}
\usepackage{subcaption}
\newcommand{\ASS}{AS2}
\newcommand{\TANDA}{T{\sc and}A}

\usepackage{soul} 

\title{ {\TANDA}: Transfer and Adapt Pre-Trained Transformer Models\\ for Answer Sentence Selection}
\author{
Siddhant Garg \thanks{Work done while the author was an intern at Amazon Alexa.} \\ University of Wisconsin-Madison\\ Madison, WI, USA \\ \texttt{sidgarg@cs.wisc.edu} \And Thuy Vu \\ Amazon Alexa\\ Manhattan Beach, CA, USA \\\texttt{thuyvu@amazon.com} \And Alessandro Moschitti \\
Amazon Alexa\\ Manhattan Beach, CA, USA \\\texttt{amosch@amazon.com}
}
 
\begin{document}

\maketitle

\definecolor{mygreen}{rgb}{0.0, 0.44, 0.0}
\newcommand{\delete}[1]{{\color{red}\st{#1}}} 
\newcommand{\touch}[1]{{\color{blue}{#1}}} 
\newcommand{\insrt}[1]{{\color{mygreen}{#1}}} 

\begin{abstract}

We propose {\TANDA}, an effective technique for fine-tuning pre-trained Transformer models for natural language tasks. Specifically, we first \emph{transfer} a pre-trained model into a model for a general task by fine-tuning it with a large and high-quality dataset. We then perform a second fine-tuning step to \emph{adapt} the transferred model to the target domain. We demonstrate the benefits of our approach for answer sentence selection, which is a well-known inference task in Question Answering. We built a large scale dataset to enable the transfer step, exploiting the Natural Questions dataset.
Our approach establishes the state of the art on two well-known benchmarks, WikiQA and TREC-QA, achieving MAP scores of 92\% and 94.3\%, respectively, which largely outperform the previous highest scores of 83.4\% and 87.5\%, obtained in very recent work.
We empirically show that {\TANDA} generates more stable and robust models reducing the effort required for selecting optimal hyper-parameters.
Additionally, we show that the transfer step of {\TANDA} makes the adaptation step more robust to noise.
This enables a more effective use of noisy datasets for fine-tuning. Finally, we also confirm the positive impact of {\TANDA} in an industrial setting, using domain specific datasets subject to different types of noise.
\end{abstract}

\section{Introduction}
In recent years, virtual assistants have become a central asset for technological companies.
This has increased the interest of AI researchers in studying and developing conversational agents, some popular examples being Google Home, Siri and Alexa. This has renewed the research interest in Question Answering (QA) and, in particular, in two main tasks: (i) answer sentence selection (AS2), which, given a question and a set of answer sentence candidates, consists in selecting sentences (e.g., retrieved by a search engine) correctly answering the question; and (ii) machine reading (MR) \cite{DBLP:journals/corr/ChenFWB17} or reading comprehension, which, given a question and a reference text, consists in finding a text span answering it. Even though the latter is gaining more and more popularity, AS2 is more relevant to a production scenario since, a combination of a search engine and an AS2 model already implements an initial QA system.

The AS2 task was originally defined in the TREC competition \cite{wang-etal-2007-jeopardy}. 
With the advent of neural models, it has had significant contributions from techniques such as  \cite{he-lin-2016-pairwise,DBLP:journals/corr/abs-1801-01641,DBLP:journals/corr/WangJ16b}.

Recently, approaches for training neural language models, e.g., ELMO \cite{DBLP:journals/corr/abs-1802-05365}, GPT \cite{noauthororeditor}, BERT \cite{DBLP:journals/corr/abs-1810-04805}, RoBERTa  \cite{DBLP:journals/corr/abs-1907-11692},  XLNet \cite{DBLP:journals/corr/abs-1901-02860} have led to major advancements in several NLP subfields. These methods capture dependencies between words and their compounds by pre-training neural networks on large amounts of data. Interestingly, the resulting models can be easily applied to solve different NLP applications by just fine-tuning them on the training data of the target tasks. 
For example, the Transformer \cite{NIPS2017_7181} can be pre-trained on a large amount of data obtaining a powerful language model, which can then be specialized for solving specific NLP tasks by just adding new layers and training them on the target data. 
Although the approach is simple, the procedure for fine-tuning a Transformer-based model is not completely understood, and can result in high accuracy variance, with a possible on-off behavior, e.g., models may always predict a single label. As a consequence, researchers need to invest a considerable effort in selecting suitable parameters, with no theory or a well-assessed best practice helping them. Such problem also affects QA, and, in particular, AS2 since no large and and accurate dataset has been developed for it.

In this paper, we study the use of Transformer-based models for AS2 and provide effective solutions to tackle the data scarceness problem for AS2 and the instability of the fine-tuning step. In detail, the contributions of our papers are:
\begin{itemize}
\item We improve stability of Transformer models by adding an intermediate fine-tuning step, which aims at specializing them to the target task (AS2), i.e., this step transfers a pre-trained language model to a model for the target task. 
\item We show that the transferred model can be effectively adapted to the target domain with a subsequent fine-tuning step, even when using target data of small size.
\item Our Transfer and Adapt ({\TANDA}) approach makes fine-tuning: (i) easier and more stable, without the need of cherry picking parameters; and (ii) robust to noise, i.e., noisy data from the target domain can be utilized to train an accurate model.
\item We built ASNQ, a dataset for AS2, by transforming the recently released Natural Questions (NQ) corpus \cite{47761} from MR to AS2 task. This was essential as our transfer step requires a large and accurate dataset. ASNQ is an important contribution of our work to the research community. \footnote{The ASNQ dataset and trained models can be accessed at \url{https://github.com/alexa/wqa_tanda}}
\item Finally, the generality of our approach and  empirical investigation suggest that our {\TANDA} findings also apply to other NLP tasks, especially, textual inference, although empirical analysis is essential to confirm these claims.
\end{itemize} 
We evaluate {\TANDA} on well-known academic benchmarks, i.e., TREC-QA \cite{wang-etal-2007-jeopardy} and WikiQA \cite{yang-etal-2015-wikiqa}, as well as three different industrial datasets, where questions are derived from Alexa Traffic and candidate sentences are selected from web data. The results show that:

\begin{itemize}
\item {\TANDA} improves the stability of Transformer models. In the adapt step, the model accuracy throughout different epochs show a smooth and convex behavior, which is ideal for estimating optimal parameters.
\item We improved the state of the art in AS2 by just applying BERT and RoBERTa to AS2 and we further improved it by almost 10 absolute percent points in MAP scores with {\TANDA}.
\item {\TANDA} achieves much higher accuracy than traditional fine-tuning, especially in case of noise data. For example, the drop in performance is up to one order of magnitude lower with {\TANDA}, i.e., ~2.5\%, when we inject 20\% of noise in the WikiQA and TREC-QA datasets.
\item Our experiments with real-world datasets built from Alexa traffic data confirm all our above findings. Specifically, we observe the same robustness to noise, which, in this case, is generated by real sources.
\end{itemize}

The rest of the paper is structured as follows: we describe related work in Section~\ref{relwork}, details of our {\TANDA} approach in Section~\ref{tanda}, present the new AS2 dataset in Section~\ref{ASNQ}. The experimental results on academic benchmarks, including experiments with noise data, are reported in Section~\ref{bench}, while the results on real-world data are illustrated in Section~\ref{alexa}. Finally, we derive conclusions in Section~\ref{con}.

\section{Related Work}
\label{relwork}
Recent AS2 models are based on Deep Neural Networks (DNNs), which learn distributed representations of the input data and are trained to apply a series of non-linear transformations to the input question and answer, represented as compositions of word or character embeddings. DNN architectures learn answer sentence-relevant patterns using intra-pair similarities as well as cross-pair, question-to-question and answer-to-answer similarities, when modeling the input texts. For example, the CNN network by \citeauthor{Severyn:2015:LRS:2766462.2767738}
has two separate embedding layers for the question and answer, which are followed by the respective convolution layers. The output of the latter is concatenated and then passed through the final fully-connected joint layer. They also added embeddings encoding relational links between matching words \cite{DBLP:journals/corr/SeverynM16}: a sort of hardcoded attention, which highly increases  accuracy.

More recent papers \cite{shen-etal-2017-inter,tran-etal-2018-context,DBLP:journals/corr/abs-1806-00778} also propose a technique of inter-weighted alignment networks for this task. While others \cite{DBLP:journals/corr/WangJ16b,Bian:2017:CMD:3132847.3133089,DBLP:journals/corr/abs-1905-12897} use a compare aggregate architecture, which also exploits an attention mechanism over the question and answer sentence embeddings. \citeauthor{tayyar-madabushi-etal-2018-integrating} (2018) propose a method that integrates fine-grained Question Classification with a Deep Learning model designed for {\ASS}.

Previous work \cite{wang-etal-2007-jeopardy} was carried out on relatively small datasets compared to other NLP tasks, such as machine translation, e.g., the WMT15 English-Czech dataset \cite{luong2016acl_hybrid} contains 15.8 million sentence pairs. This further motivates our work of creating a new large scale AS2 dataset, ASNQ, which is two orders of magnitudes larger than datasets such as TREC-QA \cite{wang-etal-2007-jeopardy}. 

ASNQ is based on the NQ dataset, which is a corpus designed to (i) study MR tasks, and (ii)
solve several problems of previous corpora such as SQuAD \cite{DBLP:journals/corr/RajpurkarZLL16}. 
MR models, e.g., \cite{DBLP:journals/corr/SeoKFH16}, are different from those used for AS2 and beyond the purpose of our paper. It is still worthwhile to mention that \citeauthor{min-etal-2017-question} (2017) explored transfer learning for the BiDAF model for MR.

Very recent works \cite{DBLP:journals/corr/abs-1810-04805,yang-etal-2019-end-end} use pre-trained Transformer models for MR. In this context, {\TANDA} and ASNQ may provide alternative research directions, e.g., a more stable model that can then be fined tuned to the MR task. 

\citeauthor{Wang2019ToTO} (2019) report some marginal improvement for the task of text classification by fixing weights of Transformer models derived by BERT, when training the classification layer. \citeauthor{DBLP:journals/corr/abs-1905-05583} (2019) carried out additional pre-training of BERT-derived models on the target dataset. They also used different learning rates for different model layers.

\begin{figure*}[t]
    \includegraphics[width=\textwidth]{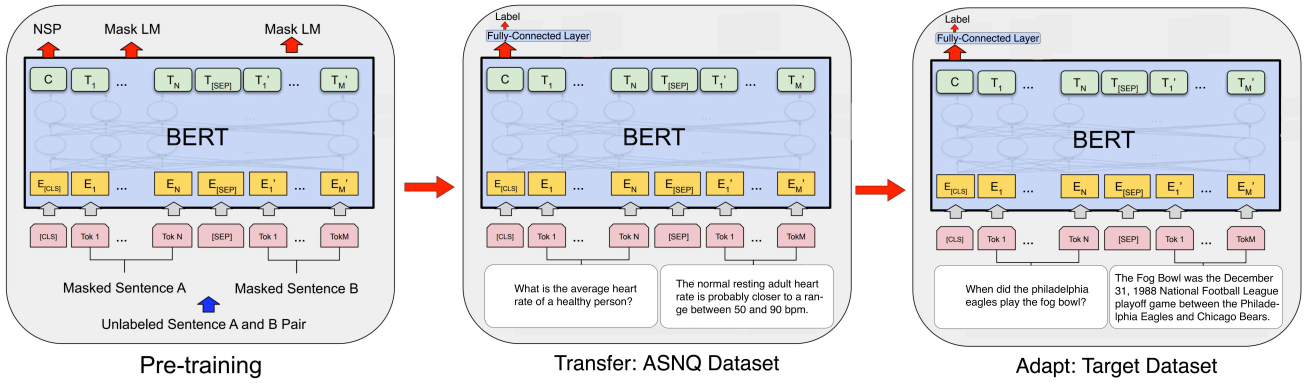}
\vspace{-1em}
    \caption{Transfer and Adapt for Answer Sentence Selection, applied to BERT}
    \label{fig:tanda}
\vspace{-.5em}
\end{figure*}

\section{{\TANDA}: Transfer and Adapt}
\label{tanda}

We propose to train Transformer models for the AS2 by applying a two-step fine-tuning, called Transfer {\sc and} Adapt ({\TANDA}). The first step transfers the language model of the Transformer to the AS2 task; and the second fine-tuning step adapts the obtained model to the specific target domain, i.e., specific types of questions and answers. We first provide a background on AS2 and Transformer models, and, then explain our approach in more detail.

\subsection{AS2 task and model definition}
AS2 can be defined as follows: given a question $q$ and a set of answer sentence candidates $S=\{s_1,..,s_n\}$, select a sentence $s_k$ that correctly answers $q$.
We can model the task as a function $r: Q \times  \mathcal{P}(S) \rightarrow S$, defined as $r(q,S) = s_k$, where $k={\tt argmax}_i \hspace{.3em} p(q,s_i)$ and $p(q,s_i)$ is the probability of correctness of $s_i$. We estimate $p(q,s_i)$ using neural networks, in particular, Transformer models, as explained below.


 \begin{figure}[h]
   \center
    \includegraphics[width=0.35\textwidth]{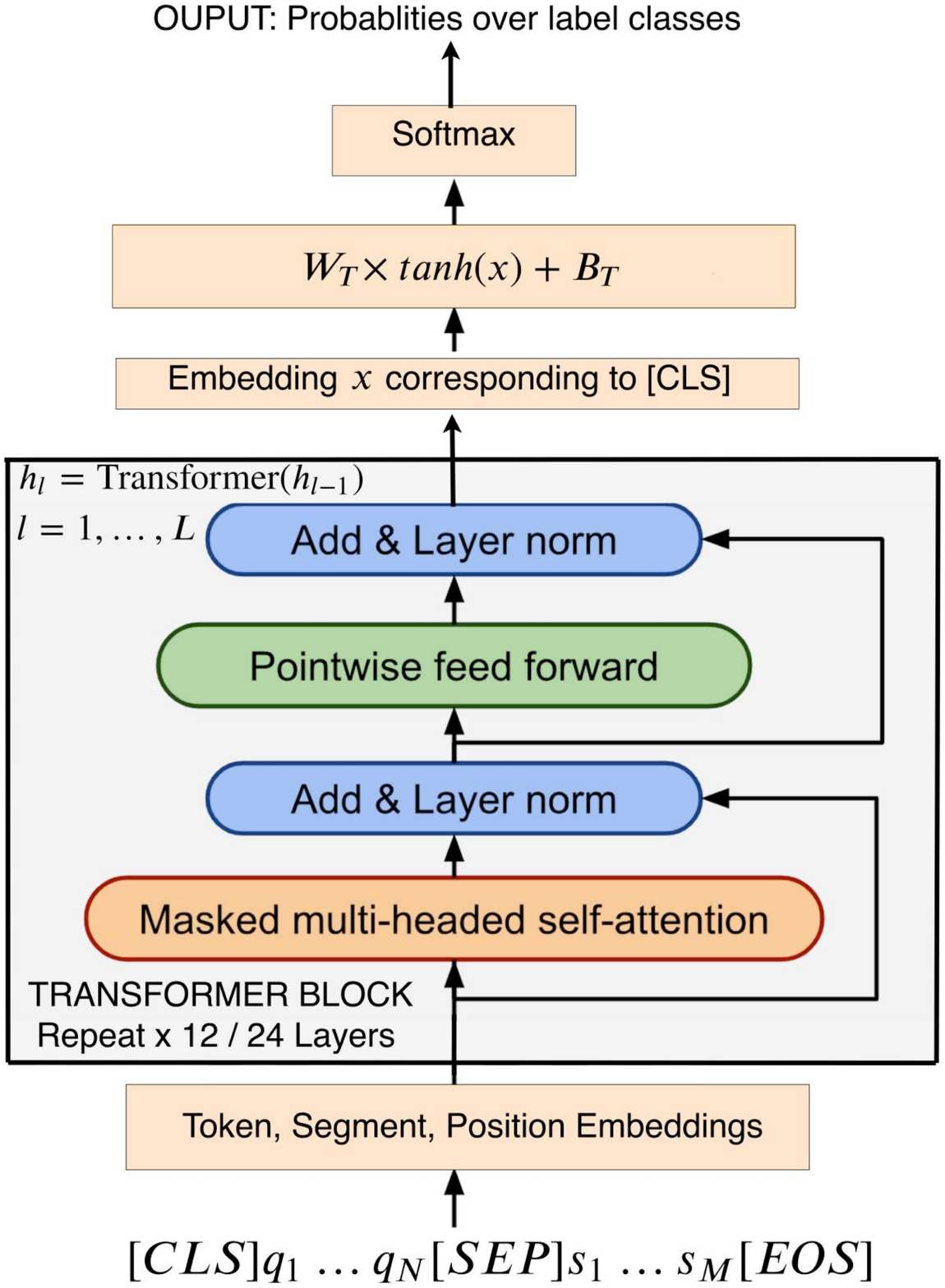}
\vspace{-.5em}
    \caption{Transformer architecture with on top a linear classifier for fine-tuning on AS2. Here [CLS] Tok$^1_1$,...,Tok$^1_N$ [SEP] Tok$^2_1$,...,Tok$^2_M$[EOS] is the input to the model.}
    \label{fig:transformer}
\vspace{-.5em}
\end{figure}

\subsection{Transformers for AS2}


Transformer-based models are neural networks designed to capture dependencies between words, i.e., their interdependent contexts. Fig.~\ref{fig:transformer} shows the standard architecture for a text pair classification task. The input consists of two pieces of text, Tok$^1_1$,...,Tok$^1_N$ and Tok$^2_1$,...,Tok$^2_M$ delimited by three tags, [CLS], [SEP] and [EOS] (beginning, separator and end of sentence). The input is encoded as embeddings based on tokens, segments and their positions. These are fed as input to several blocks (up to 24) containing layers for multi-head attention, normalization and feed forward processing. The result of this transformation is an embedding, $\mathbf{x}$, representing the text pair, which models the dependencies between words and segments of the two sentences.
For a downstream task, $\mathbf{x}$ is fed (after applying a non linearity function) to a fully connected layer having weights: $W_T$ and $B_{T}$. The output layer can be used to implement the task function. For example, a softmax can be used to model the probability of a text pair classification, as described by the equation:
$\hat{y} = softmax(W_{T} \times tanh(x) + B_{T})$. 

Theoretically, this model can be trained using the log cross-entropy loss: $\mathcal{L}=-\sum_{l \in \{0,1\}} y_l \times log(\hat{y}_l)$ on pairs of text. In practice, a large amount of supervised data will be required for this step of training. The important contribution by Devlin et al., \shortcite{DBLP:journals/corr/abs-1810-04805} was to pre-train the language model, i.e., the sentence pair representation, on using surrogate tasks such as masked language modeling and next sentence prediction.

The left block in Fig.~\ref{fig:tanda} illustrates the pre-training step of a Transformer model, highlighting that some words in the pair are masked. This way the model learns to generalize the input while providing the same output. 
The default transfer approach, defined in previous work, fine-tunes the Transformer to the target task and domain (in one shot). For AS2, the training data comprises of question and sentence pairs with positive or negative labels according to the test: the sentence correctly answers the question or not.  This fine-tuning is rather critical as the initial task learned during the pre-training stage is very different from AS2. 

When only small target data is available, the transfer process from the language model to AS2 task is unstable. We conjecture that a large number of examples are required to fine-tune the large number of Transformer parameters on the new task. An evidence of this is the on-off effect, that is, the fine-tuned model always predicts a single label for all examples. More in general, the model accuracy is unstable showing a large variance over different fine-tuning attempts. 

We explain this behavior considering the quality and quantity of the training data required for the transfer step (from the pre-trained model to AS2). More precisely, a small number of data examples prevents an effective convergence to the task, while noisy data leads to incorrect convergence. Thus, we propose to divide the fine-tuning process in two steps: transfer to the task and then adapt to the target domain ({\TANDA}). 
This is advantageous over a single fine-tuning step, since the latter would require either (i) the availability of a large dataset for the target domain, which is undesirable due to higher difficulty and cost of collection of domain specific data over general data; or (ii) merging the general and domain specific data in a single training step, which is not optimal since the the model needs to be specialized only to the target data. 
In principle when using a combination of general and domain specific data, instance weighting can be used by giving more importance to the target data. However, finding the right weights is complex as neural models do not converge to a global optimum: thereby leading to very different outcomes for different weights.

\subsection{{\TANDA}}

The two steps of our approach are depicted in the center and right blocks of Fig.~\ref{fig:tanda}. We apply a standard fine-tuning step using a large scale general purpose dataset for AS2. This step is supposed to transfer the Transformer language model to the AS2 task. The resulting model will not perform optimally on the data of the target domain due to the specificity of the latter. We thus apply a second fine-tuning step to adapt the classifier to the target AS2 domain.
For example, in the transfer step, we may have general questions such as, \emph{What is the average heart rate of a healthy person} while, in the adapt step, the target domain, e.g., sport news, may contain specific questions such as: \emph{When did the Philadelphia eagles play the fog bowl?}

Using different training steps on the target data to improve performance is a rather intuitive approach.
In this paper, we highlight the role of the transfer step, which (i) greatly reduces the amount of data required in the adaptation step; and (ii) stabilizes the model, making it robust to noise.
We empirically demonstrate both claims in our experiments.

\section{Answer-Sentence Natural Questions}
\label{ASNQ}

We needed an accurate, general and large AS2 corpus to validate the benefits of {\TANDA}. Since existing AS2 datasets are small in size, we built a new AS2 dataset called Answer Sentence Natural Questions (ASNQ) derived from the recently released Google Natural Questions (NQ) dataset~\cite{47761}.

NQ is a large scale dataset intended for the MR task, where each question is associated with a Wikipedia page. For each question, a long paragraph (\texttt{long\_answer}) that contains the answer is extracted from the reference page. Each \texttt{long\_answer} may contain phrases annotated as \texttt{short\_answer}. A \texttt{long\_answer} can contain multiple sentences, thus NQ is not directly applicable for AS2.

For each question in ASNQ, the positive candidate answers are those sentences that occur in the long answer paragraphs in NQ and contain \emph{annotated} short answers. The remaining sentences from the document are labeled as negative for the target question. The negative examples can be of the following types:
\begin{enumerate}
\item Sentences from the document that are in the \texttt{long\_answer} but do not contain the annotated short answers. It is possible that these sentences might contain the \texttt{short\_answer}.
\item Sentences from the document that are not in the \texttt{long\_answer} but contain the \texttt{short\_answer} string, that is, such occurrence is purely accidental.
\item Sentences from the document that are neither in the \texttt{long\_answer} nor contain the \texttt{short\_answer}.
\end{enumerate}
\noindent 

\begin{figure}[t]
   \center
    \includegraphics[width=\linewidth]{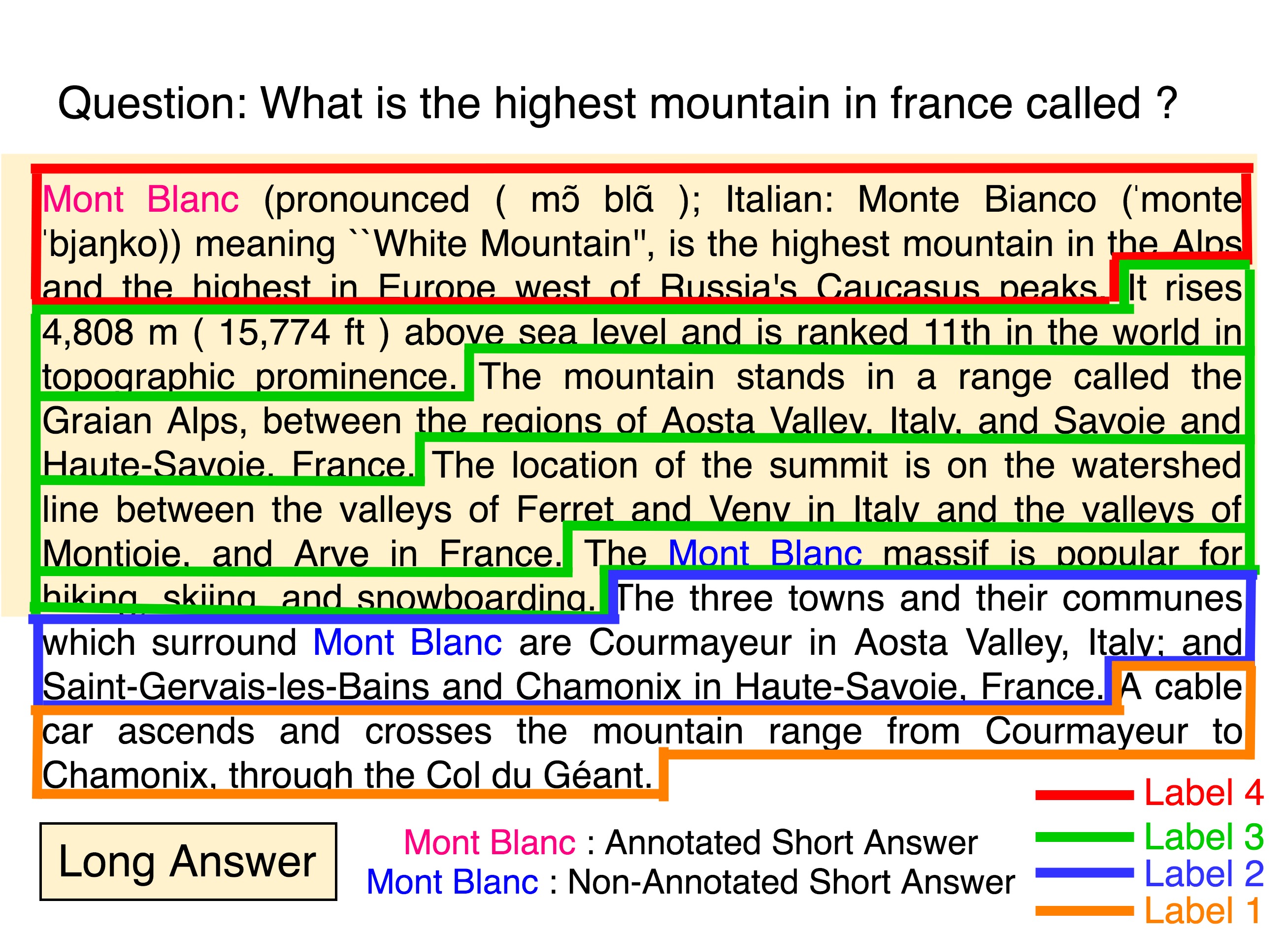}
    \caption{An example of data instance conversion from NQ to ASNQ.}
    \label{ASNQ_exp}
\end{figure}

The generation of negative examples is particularly impactful to the robustness of the model in identifying the best answer out of the similar but incorrect ones.  ASNQ has four labels that describe possible confusing levels of a sentence candidate.
We apply the same processing both to training and development sets of NQ. An example is shown in Fig.~\ref{ASNQ_exp}, while the ASNQ statistics are reported in Table~\ref{Table:ASNQ_labels}.
\begin{table}
\small
\center
\begin{center}
 \begin{tabular}{|c | c | c | c | c |} 
 \hline
 Label & S $\in$ LA & SA $\in$ S & \# Train & \# Dev  \\
 \hline
  1 & No & No & 19,446,120 & 870,404  \\ 
 \hline
  2 & No & Yes & 428,122 &  25,814\\
 \hline
  3 & Yes & No & 442,140 & 29,558\\
 \hline
  4 & Yes & Yes & 61,186 &  4,286 \\
 \hline
\end{tabular}
\caption{Label description for ASNQ. 
Here S, LA, SA refer to answer sentence, long answer passage and short answer phrase respectively.}
\label{Table:ASNQ_labels}
\end{center}
    \vspace{-1em}
\end{table}

ASNQ contains 57,242 distinct questions in the training set and 2,672 distinct questions in the dev.~set, which is an order of magnitude larger than most public AS2 datasets. For the transfer step in {\TANDA},  we use ASNQ sentence pairs with labels 1,2 and 3 as negatives and label 4 as positives.

\section{Experiments on Standard Benchmarks}
\label{bench}
We provide empirical evidence on the benefits of using {\TANDA} on two commonly used benchmarks for AS2: \mbox{WikiQA} and TREC-QA, which enable a direct comparison with previous work.

\subsection{Academic Datasets}
\textbf{WikiQA} is an AS2 dataset \cite{yang-etal-2015-wikiqa} constituted by questions from Bing query logs and candidate answer sentences extracted from Wikipedia, which are then manually labeled. Some questions have no correct answer sentence (\textit{all-}) or have only correct answer sentences \mbox{(\textit{all+})}. Table \ref{Table:WikiQA_description}  reports the corpus statistics of the versions: \emph{raw} (as distributed), without \emph{all-} questions, and without both \textit{all-} and \textit{all+} questions (clean). We follow the most used setting: training with the \emph{no all-} mode and then answer candidate sentences per question in testing with the \emph{clean} mode.

\begin{table}[t]
\small
\center
\begin{tabular}{|l|l|l|l|l|l|l|}
\hline
\multirow{2}{*}{Mode} & \multicolumn{2}{l|}{Train} & \multicolumn{2}{l|}{Dev} & \multicolumn{2}{l|}{Test} \\ \cline{2-7} 
                      & Q           & A            & Q          & A           & Q           & A           \\ \hline
raw                   & 2118    & 20360        & 296        & 2733        & 633         & 6165        \\ \hline
no all-              & 873         & 8672         & 126        & 1130        & 243         & 2351        \\ \hline
clean                 & 857         & 8651         & 121        & 1126        & 237         & 2341        \\ \hline
\end{tabular}
\caption{WikiQA dataset statistics}
\label{Table:WikiQA_description}
\end{table}

\noindent \textbf{TREC-QA} is another popular benchmark \cite{wang-etal-2007-jeopardy} for {\ASS}. We use the same splits of the original data, using the larger provided training set (TRAIN-ALL). This is noisier data, which, on the other hand, gives us more examples for training. We removed  questions without answers, or with only correct or only incorrect answer sentence candidates, from the dev.~and test sets. This resulted in $1,229$, $65$ and $68$ questions, and $53,417$, $1,117$ and $1,442$ question-answer pairs for training, dev.~and test sets, respectively. This setting refers to the \emph{Clean} setting \cite{shen-etal-2017-inter}, which is a TREC-QA standard. 

\noindent \textbf{QNLI}, namely, Question Natural Language Inference
is a dataset \cite{DBLP:journals/corr/abs-1804-07461} derived from the Stanford Question Answering Dataset (SQuAD) \cite{DBLP:journals/corr/RajpurkarZLL16} by converting question-paragraph pairs into sentence pairs, resulting in dataset with 105k question-answer pairs for training and 5.4k pairs in the dev.~data.
We carry out experiments  of using QNLI for the transfer step in {\TANDA} and compare it with previous methods \cite{DBLP:journals/corr/abs-1905-12897}, which are based on QNLI for sequential fine-tuning of non-Transformer models for AS2.  

\subsection{Training and testing details}
\subsubsection{Metrics} We measure system accuracy with Mean Average Precision (MAP) and Mean Reciprocal Recall (MRR) evaluated on the test set, using the entire set of candidates for each questions (this varies according to the different datasets). 

\subsubsection{Models} We use the pre-trained BERT-Base (12 layer), BERT-Large (24 layer), RoBERTa-Base (12 layer) and RoBERTa-Large-MNLI (24 layer) models, which were released as checkpoints for use in downstream tasks. 

\begin{table}[t]
\small
\center
\begin{tabular}{|l|l|l|}
\hline
\textbf{Model}                                                                         & \textbf{MAP}   & \textbf{MRR}   \\ \hline
Comp-Agg + LM + LC                                                            & 0.764 & 0.784 \\ \hline
Comp-Agg + LM + LC+ TL(QNLI)                                                  & 0.834 & 0.848 \\ \hline
\hline
BERT-B	FT WikiQA                                                           & 0.813 & 0.828 \\ \hline
BERT-B	 FT ASNQ                                                             & 0.884 & 0.898 \\ \hline
BERT-B	  {\TANDA} (ASNQ  $\rightarrow$ WikiQA )                                                 & 0.893 & 0.903 \\ \hline \hline
BERT-L	 FT WikiQA                                                          & 0.836 & 0.853 \\ \hline
BERT-L	 FT ASNQ                                                            & 0.892 & 0.904 \\ \hline
BERT-L	  {\TANDA} (ASNQ  $\rightarrow$ WikiQA)                                                 & 0.904 & 0.912 \\ \hline \hline
RoBERTa-B	 FT ASNQ                                                             & 0.882 & 0.894 \\ \hline
RoBERTa-B	  {\TANDA} (ASNQ  $\rightarrow$ WikiQA)                                                  & 0.889 & 0.901 \\ \hline
RoBERTa-L	 FT ASNQ                                                            & 0.910 & 0.919 \\ \hline
RoBERTa-L	  {\TANDA} (ASNQ  $\rightarrow$ WikiQA )                                                & \textbf{0.920} & \textbf{0.933} \\ \hline
\end{tabular}%
\caption{Performance of different models on WikiQA dataset. 
Here Comp-Agg + LM + LC refers to a Compare-Aggregate model with Language Modeling and Latent Clustering as proposed by Yoon et al. \shortcite{DBLP:journals/corr/abs-1905-12897}. 
TL(QNLI) refers to Transfer Learning from the QNLI corpus. L and B stand for Large and Base, respectively.}
\label{Table:WikiQA_results}
\vspace{-0.5em}
\end{table}


\subsubsection{Training} 
We adopt Adam optimizer \cite{Kingma2014AdamAM} with a learning rate of 2e-5 for the transfer step on the ASNQ dataset and a learning rate of 1e-6 for the adapt step on the target dataset. 
We apply early stopping on the dev. set of the target corpus for both steps based on the highest MAP score. We set the max number of epochs equal to 3 and 9 for adapt and transfer steps, respectively.
We set the maximum sequence length for BERT/RoBERTa to 128 tokens.

\subsubsection{Parameter Tuning} 
We selected learning rates for Adam optimizer in the transfer and adapt steps as follows: (i) We tested a \emph{reasonable} set of values for the transfer and adapt steps, identifying two promising values, $1e-6$ and $2e-5$, for the former, and five values $\{1e-6, 2e-6, 5e-6, 1e-5, 2e-5\}$ for the latter. These are within the range of typical learning rates for the Adam optimizer. (ii) We tested the ten combinations for the {\TANDA} approach and we selected the pair value, ($2e-5$,$1e-6$) corresponding to the transfer and adapt step respectively, which optimizes the MAP on the dev.~set of the target dataset. As {\TANDA} makes fine-tuning stable, we end up selecting the same parameter values for all tested datasets. It should be noted that the optimality of a larger learning rate for the first step and a smaller learning rate for the second step supports our claim of considering the second step of {\TANDA} as domain adaptation process.
For the experiments with one fine-tuning step (baseline), we again choose values between $1e-6$ and $2e-5$, i.e, following the common practice of BERT fine-tuning. 

\begin{table}[t]
\small
\center
\begin{tabular}{|l|l|l|}
\hline
\textbf{Model}                                                                         & \textbf{MAP}   & \textbf{MRR}   \\ \hline
Comp-Agg + LM + LC                                                            & 0.868 & 0.928 \\ \hline
Comp-Agg + LM + LC + TL(QNLI)                                                 & 0.875 & 0.940 \\ \hline\hline
BERT-B	 FT TREC-QA                                                           & 0.857 & 0.937 \\ \hline
BERT-B	 FT ASNQ                                                             & 0.823 & 0.872 \\ \hline
BERT-B	 {\TANDA} (ASNQ  $\rightarrow$ TREC-QA)                                                  & 0.912 & 0.951 \\ \hline\hline
BERT-L	 FT TREC-QA                                                          & 0.904 & 0.946 \\ \hline
BERT-L	 FT ASNQ                                                            & 0.824 & 0.872 \\ \hline
BERT-L	  {\TANDA} (ASNQ  $\rightarrow$ TREC-QA )                                                & 0.912 & 0.967 \\ \hline\hline
RoBERTa-B	 FT ASNQ                                                             & 0.849 & 0.907 \\ \hline
RoBERTa-B	  {\TANDA} (ASNQ  $\rightarrow$TREC-QA )                                                 & 0.914 & 0.952 \\ \hline
RoBERTa-L	 FT ASNQ                                                            & 0.880 & 0.928 \\ \hline
RoBERTa-L	  {\TANDA} (ASNQ  $\rightarrow$ TREC-QA)                                                & \textbf{0.943} & \textbf{0.974} \\ \hline
\end{tabular}%
\caption{Performance of different models on TREC-QA dataset. 
Here Comp-Agg + LM + LC refers to a Compare-Aggregate model with Language Modeling and Latent Clustering as proposed in \cite{DBLP:journals/corr/abs-1905-12897}. 
TL(QNLI) refers to Transfer Learning from the QNLI corpus. L and B stand for Large and Base, respectively.}
\label{Table:TrecQA_results}
\vspace{-1.0em}
\end{table}

\subsection{Main Results}

\subsubsection{WikiQA}

Table \ref{Table:WikiQA_results} reports the MAP and MRR of different pre-trained transformers models for two methods: standard fine-tuning (FT) and {\TANDA}. The latter takes two arguments that we indicate as \textit{transfer dataset} $\rightarrow$ \textit{adapt dataset}.  The table shows that:
\begin{itemize}
\item BERT-Large and BERT-Base with standard fine-tuning on WikiQA match the current state of the art by Yoon et al.~\shortcite{DBLP:journals/corr/abs-1905-12897}.\footnote{\url{https://aclweb.org/aclwiki/Question_Answering_(State_of_the_art)}} 
The latter uses the compare aggregate model with latent clustering, ELMO embeddings, and transfer learning from the QNLI corpus.
\item  {\TANDA} provides a large improvement over the state of the art, which has been regularly contributed to by hundreds of researchers.
\item RoBERTa-Large {\TANDA} using ASNQ $\rightarrow$ WikiQA establish an impressive new state of the art for AS2 on \mbox{WikiQA} of 0.920 and 0.933 in MAP and MRR, respectively. 
\item Finally, we note that the standard fine-tuning on ASNQ already outperforms the previous state of the art. This is mainly due to the fact that as ASNQ and WikiQA are both based on answers from Wikipedia.
\end{itemize}
The next section confirms the results above on TREC-QA.


\subsubsection{TREC-QA}
Table~\ref{Table:TrecQA_results} reports the results of our experiments with TREC-QA. We note that:

\begin{itemize}
\item RoBERTa-Large {\TANDA} with ASNQ $\rightarrow$ TREC-QA again establishes an impressive performance of  0.943 in MAP and 0.974 in MRR, outperforming the previous state of the art by Yoon et al.~\shortcite{DBLP:journals/corr/abs-1905-12897}.
\item Both BERT-Base and Large fine purely tuned on the TREC-QA corpus can surpass the previous state of the art, probably because the size of TREC-QA training corpus is larger than that of WikiQA.
\item {\TANDA} improves all the models: BERT-Base, RoBERTa-Base, BERT-Large and RoBERTa-Large, outperforming the previous state of the art with all of them.
\item Finally, the model obtained with FT on just ASNQ produces the expected results: it performs much lower than any {\TANDA} model and also lower than FT on just \mbox{TREC-QA} since the target domain of TREC questions is significantly different from that of ASNQ. 
\end{itemize}

We also tried FT on the merged ASNQ and TREC-QA dataset to show that the sequential FT, i.e., {\TANDA}, improves over simply using all the data combined. BERT-Base model fine-tuned on ASNQ $\cup$ TREC-QA achieves a MAP and MRR of 0.898 and 0.929, respectively. These are significantly lower than 0.912 MAP and 0.951 MRR, obtained with {\TANDA}. We also stress the other important benefits of {\TANDA}: (i) it enables modularity, thereby avoiding to retrain on the large ASNQ data (FT on any target dataset can start from the model transferred with ASNQ); and (ii) it has a higher training efficiency (target datasets are much smaller than ASNQ).

\subsection{Properties of {\TANDA}}

\begin{figure}[t]
    \includegraphics[width=\linewidth]{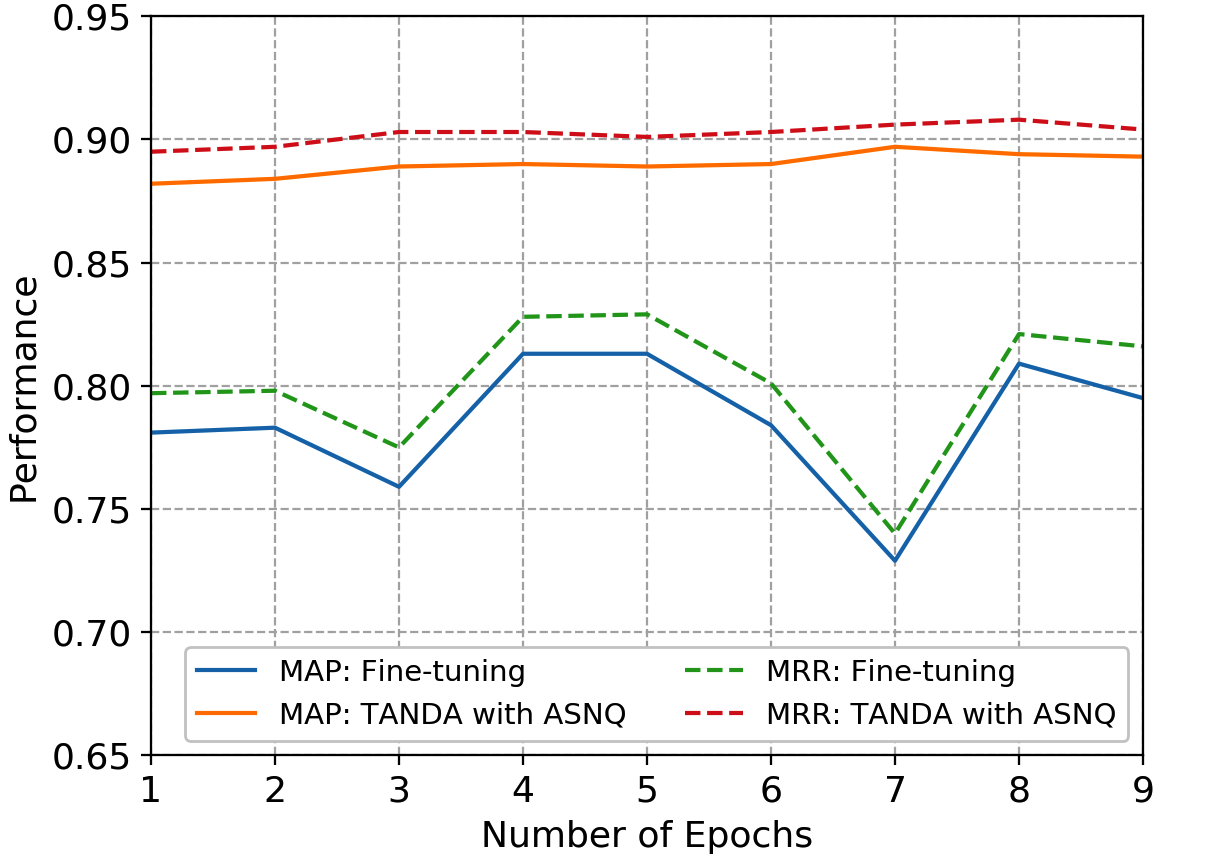}
    \caption{MAP and MRR on the WikiQA-Test-data varying with number of fine-tuning epochs on the WikiQA-Train-data for simple FT and {\TANDA}.}
\label{fig:stability_WikiQA}
\vspace{-0.5em}
\end{figure}

\begin{table*}[t]
\begin{small}
\center
\begin{tabular}{|c|c|c|c|c|c|c|c|c|}
\hline
\multirow{2}{*}{\textbf{BERT-base}} & \multicolumn{4}{c|}{WikiQA}       & \multicolumn{4}{c|}{TREC-QA}       \\ \cline{2-9} 
                       & MAP   & \% Drop & MRR   & \% Drop & MAP   & \% Drop & MRR   & \% Drop \\ \hline
No noise Fine-tuning   & 0.813 & -       & 0.828 & -       & 0.857 & -       & 0.937 & -       \\ \hline
10\% noise Fine-tuning & 0.775 & 4.67\%  & 0.793 & 4.22\%  & 0.826 & 3.62\%  & 0.902 & 3.73\%  \\ \hline
20\% noise Fine-tuning & 0.629 & \textbf{22.63}\% & 0.645 & 22.10\% & 0.738 & \textbf{13.88}\% & 0.843 & 10.03\% \\ \hline
No noise  {\TANDA} (ASNQ  $\rightarrow$ *)         & 0.893 & -       & 0.903 & -       & 0.912 & -       & 0.951 & -       \\ \hline
10\% noise  {\TANDA} (ASNQ  $\rightarrow$ *)       & 0.876 & 1.90\%  & 0.889 & 1.55\%  & 0.896 & 1.75\%  & 0.941 & 1.05\%  \\ \hline
20\% noise  {\TANDA} (ASNQ  $\rightarrow$ *)        & 0.870 & \textbf{2.57}\%  & 0.886 & 1.88\%  & 0.891 & \textbf{2.30}\%  & 0.937 & 1.47\%  \\ \hline
\end{tabular}
\caption{Model accuracy when noise is injected into WikiQA and TREC-QA datasets. $*$ indicates the target dataset for the second step of fine-tuning (adapt step).}
\label{Table:Noise_Wiki_Trec_description}
\end{small}
\vspace{-1em}
\end{table*}

\subsubsection{Stability of  {\TANDA}}
Systematic and effective fine-tuning of Transformer models is still an open problem. There is no theory or even a well-assessed best practice suggesting the optimal number of epochs to be used for fine-tuning. We claim that {\TANDA} can robustly transfer language models to the target task and this produces more stable models. For stability, we mean a low variance of the model accuracy (i) between two consecutive training epochs, and (ii) between two pairs of models that have \emph{close} accuracy on the development set. For example, BERT FT has a high variance in accuracy with the number of epochs, leading to some extreme cases of an on-off behavior, i.e, the classifier may only predict negative labels for target task (due to unbalanced datasets).

To test our hypothesis, we compared {\TANDA} with standard FT by varying the number of training epochs for the adaptation step (here we do not use early stopping for {\TANDA}).  Figure~\ref{fig:stability_WikiQA} shows the plots of MAP and MRR scores with BERT-Base on WikiQA test set with different number of training epochs on the train set. As expected FT has a high variance while {\TANDA} shows a small variance. A direct consequence of this better behavior is the higher probability to select an optimal epoch parameter on the dev.~set.

\subsubsection{Robustness to Noise in WikiQA and TREC-QA}
Better model stability also means robustness to noise. We empirically studied this conjecture by artificially injecting noise in the training sets of WikiQA and TREC-QA, by randomly sampling questions-answer pairs from the training set and switching their label values. 
We chose random samples of $10\%$ and $20\%$ of the training data, generating 867 and 1734 noisy labels, on WikiQA, respectively, and 5341 and 10683 noisy labels, on TREC-QA, respectively. We used the same \emph{Clean} data setting to directly compare with the original performance.  Table~\ref{Table:Noise_Wiki_Trec_description} shows the MAP and MRR of BERT-Base using FT and {\TANDA}, also indicating the drop percentage ($\%$ ) in accuracy due to the injection of noise. We note that standard FT is highly affected by noisy data, e.g., on \mbox{WikiQA} the accuracy decreases by 22.63\%, when 20\% of noise is injected. Very interestingly, the models using {\TANDA} are affected an order of magnitude less, i.e., just 2.57\%. A similar trend can be observed aslo for the results on TREC-QA.


\begin{table}[h]
\small
\center
\begin{tabular}{|l|l|l|l|l|}
\hline
\multirow{2}{*}{Model}    & \multicolumn{2}{l|}{WikiQA} & \multicolumn{2}{l|}{TREC-QA} \\ \cline{2-5} 
                         & MAP          & MRR          & MAP          & MRR          \\ \hline
Neg: 1 Pos: 4   					& 0.870 & 0.880 & 0.808 & 0.847         \\ \hline
Neg: 2 Pos: 4         				& 0.751 & 0.763 & 0.662 & 0.751  \\ \hline
Neg: 3 Pos: 4 						& 0.881 & 0.895 & 0.821 & 0.869  \\ \hline
Neg: 2,3 Pos: 4 					& 0.883 & 0.898 & 0.823 & 0.871  \\ \hline
Neg: 1,2,3 Pos: 4 					& 0.884 & 0.898 & 0.823  & 0.872 \\ \hline
\end{tabular}%
\caption{Impact of different labels of ASNQ on fine-tuning BERT for answer sentence selection. Neg and Pos refers to question-answer (QA) pairs of that particular label being chosen for fine-tuning.}
\label{Table:ASNQ_expts}
\vspace{-.5em}
\end{table}

\subsection{Insights on ASNQ}

\subsubsection{Ablation studies} \label{ANSQ_experiments}

Fig.~\ref{ASNQ_exp} shows that we generated different types of negative examples for the AS2 task, namely, labels 1,2 and 3. We carried out experiments by fine-tuning BERT-Base on ASNQ with specific label categories assigned to the negative class. Table~\ref{Table:ASNQ_expts} shows the results: Label 3 is the most effective negative type of the three, i.e., the models only using Label 3 as negative class are just subject to a marginal drop in performance with respect to the model using all labels. Labels 2 and 3 provide the same accuracy than the three labels as the negative class.

\subsubsection{ASNQ vs.~QNLI}
Another way to empirically evaluate the impact of ASNQ is to compare it with other similar datasets, e.g., QNLI, observing the performance of the latter when used for a simple FT or in {\TANDA}. 
Table~\ref{Table:QNLI_results} shows that both  FT and {\TANDA} using ASNQ provide significantly better performance than QNLI on  the \mbox{WikiQA} dataset. 

\begin{table}[h]
\small
\center
\begin{tabular}{|l|l|l|l|l|}
\hline
\multirow{2}{*}{\textbf{BERT-Base}}    & \multicolumn{2}{l|}{\textbf{WikiQA}} & \multicolumn{2}{l|}{\textbf{TREC-QA}} \\ \cline{2-5} 
                          & MAP          & MRR          & MAP          & MRR          \\ \hline
 FT QNLI         & 0.760        & 0.766        & 0.820        & 0.890        \\ \hline
 FT ASNQ         & 0.884        & 0.898        & 0.823        & 0.872        \\ \hline
  {\TANDA}  (QNLI  $\rightarrow$)& 0.832        & 0.852        & 0.863        & 0.906        \\ \hline
  {\TANDA}  (ASNQ  $\rightarrow$) & 0.893        & 0.903        & 0.912        & 0.951        \\ \hline
\end{tabular}%
\caption{Comparison of {\TANDA}  with ASNQ and QNLI}
\label{Table:QNLI_results}
\vspace{-1em}
\end{table}

On TREC-QA dataset the results show that (i) FT on QNLI performs better than ASNQ but (ii) when {\TANDA} uses ASNQ as transfer step, the models can better adapt to TREC-QA data than when using QNLI for the same transfer type.
On one hand, this confirms the claim about the high quality of ASNQ. It is a more general and accurate AS2 dataset and is better suited for transferring the transformer language model. On the other hand, it provides some evidence that the transfer step is very important and is not just a way for initializing  weights of the adaptation step.

\section{Experiments on data from Alexa}
\label{alexa}

To show that our results can generalize well, we tested our models using four different AS2 datasets created with questions sampled from the customers' interaction with Alexa Virtual Assistant.

\begin{table*}[h]
\begin{small}
\center
\begin{tabular}{|c|c|c|c|c|c|c|c|c|c|c|c|}
\hline
\multicolumn{3}{|c|}{\multirow{2}{*}{MODEL}}                                        & \multicolumn{3}{c|}{Sample 1} & \multicolumn{3}{c|}{Sample 2} & \multicolumn{3}{c|}{Sample 3} \\ \cline{4-12} 
\multicolumn{3}{|c|}{}                                                              & Prec@1      & MAP        & MRR        & Prec@1       & MAP         & MRR         & Prec@1        & MAP          & MRR         \\ \hline
\multirow{6}{*}{BERT}    & \multirow{3}{*}{Base}  & NAD                            & 49.80       & 0.506      & 0.638      & 52.69        & 0.432       & 0.629       & 41.86         & 0.352        & 0.543       \\ \cline{3-12} 
                         &                        & ASNQ                            & 55.06       & 0.557      & 0.677      & 44.31        & 0.395       & 0.567       & 44.19         & 0.369        & 0.561       \\ \cline{3-12} 
                         &                        & TANDA (ASNQ $\rightarrow$ NAD) & 58.70        & 0.585      & 0.703      & 58.68        & 0.474       & 0.683       & 49.42         & 0.391        & 0.613       \\ \cline{2-12} 
                         & \multirow{3}{*}{Large} & NAD                            & 53.85       & 0.537      & 0.671      & 53.29        & 0.469       & 0.629       & 43.61         & 0.395        & 0.558       \\ \cline{3-12} 
                         &                        & ASNQ                            & 57.49       & 0.552      & 0.686      & 50.89        & 0.440       & 0.630       & 45.93         & 0.399        & 0.585       \\ \cline{3-12} 
                         &                        & TANDA (ASNQ $\rightarrow$ NAD) & 61.54       & 0.607      & 0.725      & 63.47        & 0.514       & 0.727       & 51.16         & 0.439        & 0.616       \\ \hline
\multirow{6}{*}{RoBERTa} & \multirow{3}{*}{Base}  & NAD                            & 59.11       & 0.563      & 0.699      & 56.29        & 0.511       & 0.670       & 48.26         & 0.430        & 0.612       \\ \cline{3-12} 
                         &                        & ASNQ                            & 58.70       & 0.587      & 0.707      & 54.50        & 0.473       & 0.656       & 45.35         & 0.437        & 0.608       \\ \cline{3-12} 
                         &                        & TANDA (ASNQ $\rightarrow$ NAD) & 65.59       & 0.623      & 0.757      & 62.87        & 0.537       & 0.714       & 56.98         & 0.473        & 0.679       \\ \cline{2-12} 
                         & \multirow{3}{*}{Large} & NAD                            & 70.81       & 0.654      & 0.796      & 63.47        & 0.581       & 0.734       & 52.91         & 0.490        & 0.651       \\ \cline{3-12} 
                         &                        & ASNQ                            &  64.37	& 0.627	& 0.750   & 59.88        & 0.526       & 0.705       & 54.65         & 0.478        & 0.674       \\ \cline{3-12} 
                         &                        & TANDA (ASNQ $\rightarrow$ NAD) & 71.26       & 0.680      & 0.805      & 74.85        & 0.625       & 0.821       & 58.14         & 0.514        & 0.699       \\ \hline
\end{tabular}
\caption{Comparison between FT and {\TANDA} on real-world datasets derived from Alexa Virtual Assistant traffic}
\label{Table:Alexa_dataset_results}
\end{small}
\vspace{-.8em}
\end{table*}

\begin{table}[h]
\small
\center
\begin{tabular}{|c|c|c|c|c|}
\hline
Dataset             & Questions & QA Pairs & Pos. & Neg. \\ \hline
Sample 1    	& 435       & 21,530    & 19,598        & 1,932        \\ \hline
Sample 2    	& 441       & 44,593    & 40,136        & 4,457        \\ \hline
Sample 3 		& 452       & 45,300    & 42,131        & 3,169        \\ \hline
\end{tabular}
\caption{Statistics of samples 1, 2  and 3 (accurate test sets)}
\label{Table:test_dataset_description}
\vspace{-1em}
\end{table}

\subsection{Datasets}
We built three test sets based on three samples of questions labelled with information intent that can be answered using unstructured text.
Questions from Sample 1 are extracted from NQ questions while questions for samples 2 and 3 are generated from Alexa users' questions. For each questions, we selected 100 sentence candidates from the top documents retrieved by a search engine: (i) for samples 1 and 2, we used an elastic search system, ingested with several web domains, ranging from Wikipedia to \emph{reference.com}, \emph{coolantarctica.com}, \emph{www.cia.gov/library}, etc. (ii) For Sample 3, we retrieved the candidate answers using a commercial search engine, to have a higher retrieval quality. 

The statistics of the three sample test sets are reported in Table \ref{Table:test_dataset_description}.
Their aim is to capture variations in terms of question sources and retrieval systems, thus providing more general results. Additionally, since the questions that do not have an annotated answer do not affect the system accuracy, the results we provide refer to a \emph{Clean} setting, i.e., questions with all positive or negative answers are removed (no \emph{\mbox{all+}} and no \emph{\mbox{all-}}).


\begin{table}[h]
 \small
 \center
 \vspace{.5em}
\begin{tabular}{|c|c|c|c|c|}
\hline
Data Split & Questions & QA Pairs & Neg. & Pos. \\ \hline
Train      & 25,226     & 134,765  & 125,779      & 8,986       \\ \hline
Dev        & 2,802      & 14,974   & 14,014       & 960         \\ \hline
\end{tabular}
\caption{Statistics of the Noisy Alexa dataset (NAD)}
\label{Table:Alexa_description}
\end{table}

We also built a noisy dataset (NAD) with a similar approach to Sample 2, with the restriction of retrieving only 10 candidates per question. This enables the cheaper annotation of a larger number of questions, which is important to build an effective  training set. Indeed, the number of questions in NAD is one order of magnitude larger than those used for the previous samples (test questions). Additionally, NAD required an increased velocity of annotations resulting in a higher error rate, which we quantify around 20-25\% (as estimated on a small sample). The  statistics of NAD are presented in Table~\ref{Table:Alexa_description}. 

\subsection{Results}
In these experiments, we used, as usual, ASNQ for the transfer step, and NAD as our target dataset for the adapt step.  Table \ref{Table:Alexa_dataset_results} reports the comparative results using simple FT on NAD (denoted simply by NAD) and tested on samples 1, 2 and 3. We note that:
\begin{itemize}
\item applying ASNQ for the transfer step always provides improvement over FT.
\item BERT Large {\TANDA} improves over BERT Base {\TANDA}  for all the three different dataset samples.
\item Using {\TANDA}  with RoBERTa produces an even higher improvement than with BERT. 
\item All these experiments using NAD for training and accurate datasets for testing, show that  {\TANDA} is robust to real-world noise of NAD as it always provides significantly large gains over simple FT. 
\end{itemize}
 
\section{Conclusions}
\label{con}
In this paper, we have presented a novel approach for fine-tuning pre-trained Transformer models and tested it on a general natural language inference task, namely, answer sentence selection (AS2). 
Our approach, {\TANDA}, performs two fine-tuning steps sequentially: (i) on a general, large and high-quality dataset, which transfers a pre-trained model to the target task; and (ii) on the target dataset to perform domain adaptation. 
The results obtained on two well-known AS2 datasets, WikiQA and TREC-QA, show an impressive improvement over the state of the art. Additionally, we extensively experimented with an industrial setting, deriving the same results and conclusions we found with the academic benchmarks.

Our research deepens the general understanding of transfer learning for Transformer models. The first step of  {\TANDA} produces an intermediate model with three main features: (i) it can be more effectively used for fine-tuning on the target NLP application, being more stable and easier to adapt to other tasks; (ii) it is robust to noise, which might affect the target domain data; and (iii) it enables modularity and efficiency, i.e., once a Transformer model is adapted to the target general task, e.g., AS2, only the adapt step is needed for each targeted domain. This is an important advantage in terms of scalability as the data of (possibly many) different target domains can be typically smaller than the dataset for the transfer step(ASNQ), thereby causing the main computation to be factorized on the initial transfer step.

We conjecture that one caveat of using simple fine-tuning on a combination of ASNQ and target data may produce an accuracy improvement similar to that obtained using {\TANDA}. However, such a combination can be tricky to optimize as the target data requires greater weighting than the more general data of ASNQ during fine-tuning the model. Our experiments with TREC-QA show that a simple union of the dataset with ASNQ is sub-optimal than sequential fine-tuning over ASNQ followed by TREC-QA. In any case, the important modular aspect of {\TANDA} will not hold in such a scenario.

Interesting future work can be devoted to address the question about the applicability and generalization of the {\TANDA} approach to other NLP tasks. In the specific context of AS2, it would be interesting to test if ASNQ can produce the same benefits for related but clearly different tasks, e.g., paraphrasing or textual entailment, where the relation between the members of text pairs are often different from those occurring between questions and answers.

\bibliographystyle{aaai}
\bibliography{aaai2020}

\end{document}